\documentclass[conference,a4paper]{IEEEtran}

\usepackage{booktabs}

\def\IEEEkeywordsname{Keywords}

\ifCLASSINFOpdf
   \usepackage[pdftex]{graphicx}
\else
\fi
\usepackage{svg}

\usepackage{amsmath}
\usepackage{amsfonts}
\usepackage[acronym]{glossaries}
\usepackage[numbers, sort]{natbib}
\usepackage[capitalise]{cleveref}

\usepackage{stfloats}
\usepackage{eso-pic}
\usepackage{url}
\AddToShipoutPictureBG*{
  \AtPageUpperLeft{%
    \put(0,-40){\raisebox{15pt}{\makebox[\paperwidth]{\begin{minipage}{21cm}\centering
      \textcolor{gray}{This article has been accepted for publication in the proceedings of IEEE Sensors 2023.} 
    \end{minipage}}}}%
  }
  \AtPageLowerLeft{%
    \raisebox{25pt}{\makebox[\paperwidth]{\begin{minipage}{21cm}\centering
      \textcolor{gray}{ \copyright 2022 IEEE.  Personal use of this material is permitted.  Permission from IEEE must be obtained for all other uses, in any current or future media, including reprinting/republishing this material for advertising or promotional purposes, creating new collective works, for resale or redistribution to servers or lists, or reuse of any copyrighted component of this work in other works.
      }
    \end{minipage}}}%
  }
}

\begin{document}
\newacronym{mpc}{MPC}{Model Predictive Control}
\newacronym{mpcc}{MPCC}{Model Predictive Contouring Controller}
\newacronym{rl}{RL}{Reinforcement Learning}
\newacronym{mlp}{MLP}{Multilayer Perceptron}
\newacronym{forl}{FoRL}{Foundations of Reinforcement Learning}
\newacronym{ml}{ML}{Machine Learning}
\newacronym{sb3}{SB3}{Stable Baselines 3}
\newacronym{sac}{SAC}{Soft Actor Critic}
\newacronym{ppo}{PPO}{Proximal Policy Optimization}
\newacronym{ai}{AI}{Artificial Intelligence}
\newacronym{nn}{NN}{Neural Network}
\newacronym{sota}{SotA}{State-of-the-Art}
\newacronym{esc}{ESC}{Electronic Speed Controller}
\newacronym{ros}{ROS}{Robot Operating System}
\newacronym{imu}{IMU}{Inertial Measurement Unit}
\newacronym{ekf}{EKF}{Extended Kalman Filter}
\newacronym{slam}{SLAM}{Simultaneous Localization And Mapping}
\newacronym{sdc}{SDC}{Self Driving Cars}
\newacronym{obc}{OBC}{On Board Computer}
\newacronym{qp}{QP}{Quadratic Programming}
\newacronym{uav}{UAV}{Unmanned Aerial Vehicles}
\newacronym{cg}{CG}{Center of Gravity}
\newacronym{em}{EM}{Expectation Maximization}
\newacronym{rms}{RMS}{Root Mean Square}
\newacronym{map}{MAP}{Model- and Acceleration-based Pursuit}
\newacronym{pd}{PD}{Proportional-Derivative}
\newacronym{lut}{LUT}{Lookup Table}
\newacronym{RGB}{RGB}{Red Green Blue}
\newacronym{HYP}{HYP}{Hyper Spectral}
\newacronym{shm}{SHM}{Structural Health Monitoring}
\newacronym{iot}{IoT}{Internet of Things}
\newacronym{tn}{TN}{True Negative}
\newacronym{fp}{FP}{False Positive}
\newacronym{fn}{FN}{False Negative}
\newacronym{mcc}{MCC}{Matthews Correlation Coefficient}
\newacronym{pca}{PCA}{Principal Component Analysis}
\newacronym{sne}{SNE}{Stochastic Neighbor Embedding}
\newacronym{umap}{UMAP}{Uniform Manifold Approximation and Projection}
\newacronym{iot}{IoT}{Internet of Things}
\newacronym{fomo}{FOMO}{Faster Objects More Objects}
\newacronym{tflm}{TFLM}{TensorFlow Lite Micro}
\newacronym{nba}{NBA}{Normalized Bounding Box Area}
\newacronym{yolo}{YOLO}{You Only Look Once}
\newacronym{mae}{MAE}{Mean Average Error}
\newacronym{cnn}{CNN}{Convolutional Neural Network}
\newacronym{tinyml}{TinyML}{Tiny Machine Learning}
%
\title{Enhancing Lightweight Neural Networks for Small Object Detection in IoT Applications}

\author{\IEEEauthorblockN{Liam Boyle}
\IEEEauthorblockA{Dep. Information Technology \\ and Electrical Engineering\\
ETH Zürich\\
liam.boyle@pbl.ee.ethz.ch}
\and
\IEEEauthorblockN{Nicolas Baumann}
\IEEEauthorblockA{Dep. Information Technology \\ and Electrical Engineering\\
ETH Zürich\\
nicolas.baumann@pbl.ee.ethz.ch}
\and
\IEEEauthorblockN{Dr. Seonyeong Heo}
\IEEEauthorblockA{Dept. of Computer Science and Engineering\\
Kyung Hee University\\
seonyeong.heo@khu.ac.kr}
\and
\IEEEauthorblockN{Dr. Michele Magno}
\IEEEauthorblockA{Dep. Information Technology \\ and Electrical Engineering\\
ETH Zürich\\
michele.magno@pbl.ee.ethz.ch}}


\author{\IEEEauthorblockN{Liam Boyle$^{a}$, Nicolas Baumann$^{a}$, Dr. Seonyeong Heo$^{b}$, Dr. Michele Magno$^{a}$}
\IEEEauthorblockA{\textit{$^{a}$Department of Information Technology and Electrical Engineering, ETH Zurich, Z\"urich, Switzerland} \\
\IEEEauthorblockA{\textit{$^{b}$ Dept. of Computer Science and Engineering, Kyung Hee University, Yongin, South Korea}}
}}

%


\maketitle

\begin{abstract}
Advances in lightweight neural networks have revolutionized computer vision in a broad range of \gls{iot} applications, encompassing remote monitoring and process automation . However, the detection of small objects, which is crucial for many of these applications, remains an underexplored area in current computer vision research, particularly for embedded devices. To address this gap, the paper proposes a novel adaptive tiling method that can be used on top of any existing object detector including the popular \gls{fomo} network for object detection on microcontrollers. Our experimental results show that the proposed tiling method can boost the F1-score by up to 225\% while reducing the average object count error by up to 76\%. Furthermore, the findings of this work suggest that using a soft F1 loss over the popular binary cross-entropy loss can significantly reduce the negative impact of imbalanced data. Finally, we validate our approach by conducting experiments on the Sony \emph{Spresense} microcontroller, showcasing the proposed method's ability to strike a balance between detection performance, low latency, and minimal memory consumption. \newline
\end{abstract}

\begin{IEEEkeywords}
Object Detection, TinyML, IoT, Microcontrollers
\end{IEEEkeywords}


%
\IEEEpeerreviewmaketitle

\section{Introduction}

The introduction of low-power sensors has opened up many possibilities in the areas of remote monitoring, quality control, and process automation. Many of these tasks rely on accurate object detection, often of very small objects.


Today, state-of-the-art object detectors typically use \gls{cnn} \cite{yolo} to predict object locations and classes, and more recently transformer-based architectures \cite{vit} have gained considerable popularity. On the other hand, those models are extremely computationally demanding, and thus require powerful hardware that consumes several watts, making them not suitable for the majority of \gls{iot} processors \cite{eggimann2019risc}. This has prompted scholars to propose new lightweight network architectures that utilize model quantization methods and novel frameworks such as \gls{tflm} \cite{tflitemicro} to enable \gls{tinyml} on mobile and low-power embedded devices \cite{moosmann2023tinyissimoyolo, acurate_detection_edge, info_flow}.

While lightweight neural network's advances have allowed great progress to be made on tasks such as image classification \cite{conti2016accelerated}, detecting small objects remains a challenge \cite{thinNets, smallod0}, especially on embedded devices \cite{moosmann2023tinyissimoyolo}, as the computational budget on embedded processors is extremely limited \cite{magno2014low,cerutti2020sound, kartsch2018smart}. Multiple factors contribute to the difficulty of small object detection, with the foremost being that small objects are only covered by very few pixels, making it difficult to generate distinctive features. Another factor is the striding operations in CNNs which can result in small objects disappearing completely in the feature maps. Furthermore, it is worth noting that most of the publicly available datasets, such as ImageNet \cite{imagenet}, display a bias towards large objects (i.e. prominently featured objects). This, of course, adversely affects the usefulness of transfer learning which is an important tool for developing models for specific use cases with limited datasets.


\begin{figure}
\centering
\includegraphics[width=0.45\textwidth]{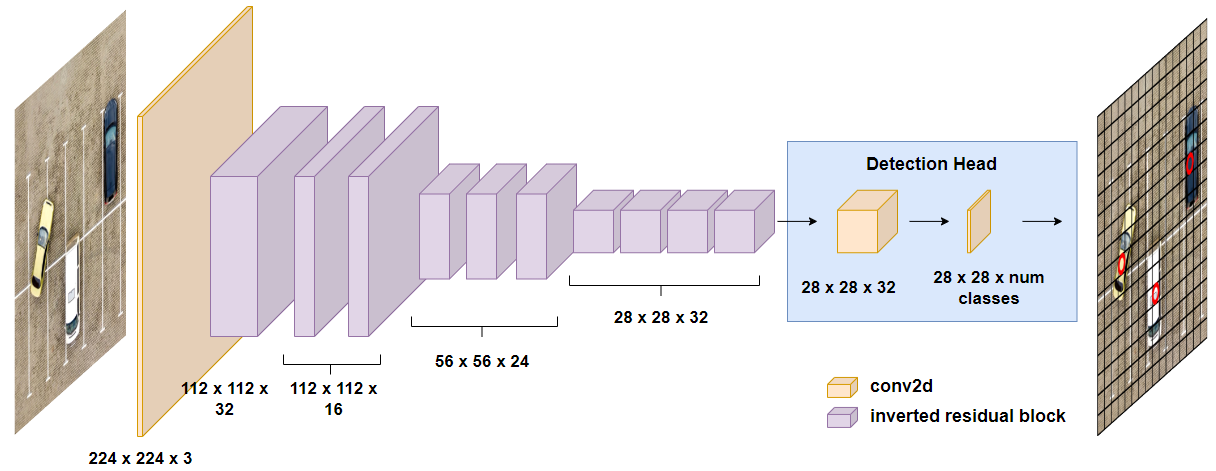}
\caption{Overview of the \gls{fomo} architecture. We show filter sizes for an example input image size of $224 \times 224 \times 3$. In this example, the feature extractor produces a $28 \times 28$ feature grid for which the detection head classifies each grid cell as containing an object or not. The final output after clustering is shown on the right.}
\label{fig:fomo-arch}
\vspace{-10pt}
\end{figure}

This work focuses on a remote monitoring field use case, namely aerial car park monitoring, and investigates the viability of lightweight embedded object detection in this scenario. This is quite challenging as it involves monitoring the scene from a significant height to provide comprehensive coverage, which in turn requires the object detector to perform well on small objects. We identify the \gls{fomo} \cite{fomo} architecture as a lightweight object detection architecture, capable of running on embedded devices, and explore its prediction abilities for our use case. While \gls{fomo} is well suited to embedded devices its network architecture is not optimized for small object detection. This paper demonstrates both the feasibility of having a quantized version running on low-power processors, in particular the Sony \emph{Spresense}, and the above-mentioned shortcomings of the original \gls{fomo} architecture. Furthermore, the paper proposes several alterations to the model that drastically improve the detection results on small objects.

The main contributions of our work are as follows:
\begin{itemize}
    \item Presentation of an adaptive tiling method to improve the signal-to-noise ratio while keeping the input resolution to the network small.
    \item We demonstrate how the soft F1 loss, introduced by Maiza \cite{softf1}, can be used to better balance precision and recall when there is a heavy imbalance between objects and background.
    \item Through experimental results, using the Sony \emph{Spresense} platform, we show how our method can be used to trade off latency for low memory consumption and good detection performance of small objects on low-power microcontrollers.
\end{itemize}

\section{Method}

\subsection{\gls{fomo}}
\gls{fomo} \cite{fomo} is a lightweight object detection network that predicts object centers as opposed to bounding boxes. Solving this simplified problem results in a much smaller network than for example YOLO \cite{yolo}, which allows \gls{fomo} to run on microcontrollers. FOMO uses the early layers of the MobileNets V2 \cite{mobilenetsv2} network as a feature extractor, resulting in an $n\times n$ feature grid, where the size of $n$ depends on how deep the feature extractor is. The feature grid is then fed into a small detection head that predicts object centers by classifying each of the $n\times n$ features and clustering predictions that belong to the same object. The original \gls{fomo} network, as implemented by \emph{Edge Impulse}, uses a crude clustering technique, whereby all predictions on neighboring grid cells are joined together. In \cref{fig:fomo-arch} we show an overview of the \gls{fomo} architecture and an example of the output that this network generates.

This architecture has two key limitations. Firstly, it only makes one prediction for each of the $n\times n$ features, unlike other single-shot detectors, such as \gls{yolo} \cite{yolo}. This means that small objects which are localized in the same grid cell will count as one prediction. Secondly, it can not distinguish between nearby objects that are predicted in neighboring grid cells because the clustering step will incorrectly fuse them together. 

\subsection{Adaptive Tiling}
\begin{figure}
\centering
\includegraphics[width=0.4\textwidth]{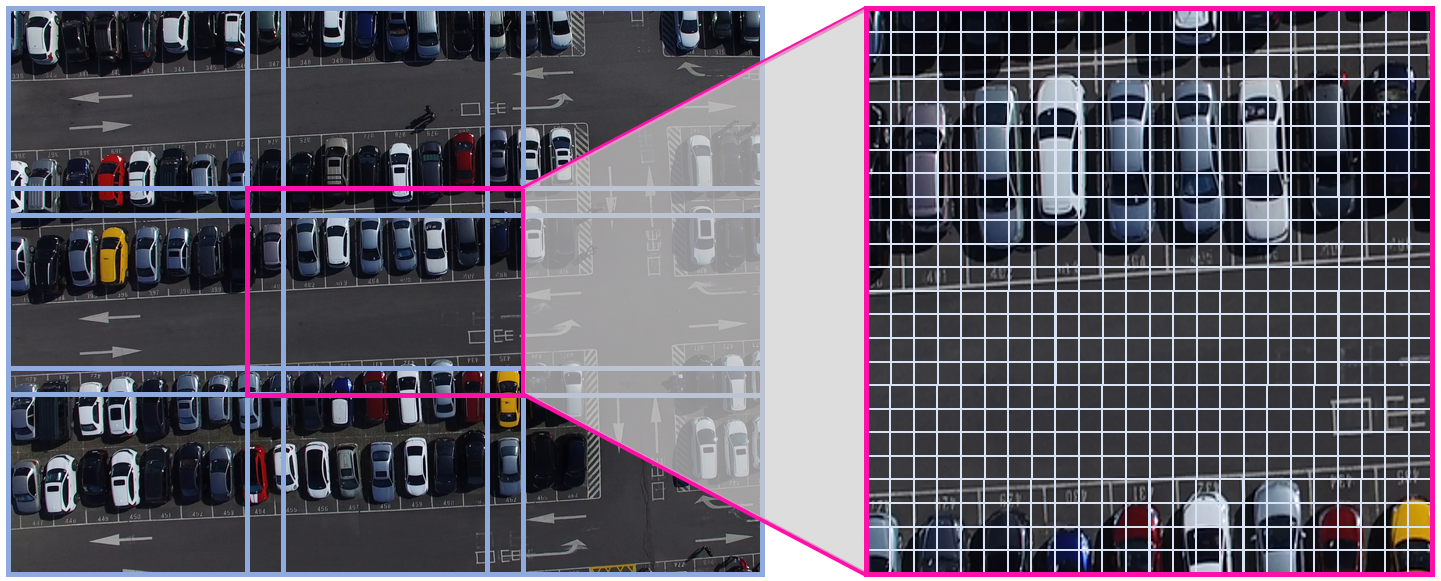}
\caption{In this example the image on the left is split into nine overlapping tiles. In the highlighted tile on the right, we overlay the feature grid resolution on which predictions will be made.}
\label{fig:tiling-sheep}
\vspace{-10pt}
\end{figure}

To address the general issue of lacking pixel information for small objects as well as the object size limitation stemming from the \gls{fomo} architecture, we propose an adaptive tiling approach to split images into smaller sub-images. Splitting images into tiles has multiple advantages. Firstly, it increases the relative size of the objects thus increasing the signal-to-noise ratio assuming that we keep the input resolution to the network constant. Secondly, increasing the size of the objects reduces the probability of multiple objects being located in the same grid cell. Furthermore, with the relative increase in size of the objects one can counteract the bias in most datasets towards larger objects.
To avoid mispredictions on objects that are on the tile border when splitting the image we follow the work of Akyon et al. \cite{sahi} and split the images into overlapping tiles. The overlap is set to be 1.5 times the average object width/height to ensure that every object is fully visible in at least one tile. \cref{fig:tiling-sheep} shows an example of how the proposed process tiles an image into smaller tiles for individual processing. This example demonstrates how the tiling increases the relative size of the objects which reduces the likelihood of multiple objects being located in the same grid cell.

The drawback of splitting images into tiles for individual processing is the resulting increase in processing time. The smaller the objects are; the more tiles need to be used to get good detection results. In this work, we propose an adaptive tiling approach that uses a target object size, which we quantify using the \gls{nba}, to adjust the number of tiles. For images taken close to the objects, only a few tiles are necessary to achieve the target object size, and vice versa for images taken far away from the objects. This prevents the system from splitting images into more tiles than necessary and thus reduces processing time. In the example shown in \cref{fig:tiling-sheep} the \gls{nba} of the full image is 0.0045, thus, to reach a target \gls{nba} of 0.04 the content of the tiles has to be about 9 times larger than in the original image. This is achieved by splitting the image into 9 overlapping tiles. During training, we use the bounding box size of our labeled objects to calculate the optimal number of tiles, and in real-world applications, one could use height measurements from GPS for example to estimate the object size. The \gls{nba} therefore serves as a prior for the object size to optimize detection performance and computation.

\begin{figure*}
\centering
\includegraphics[width=0.9\textwidth]{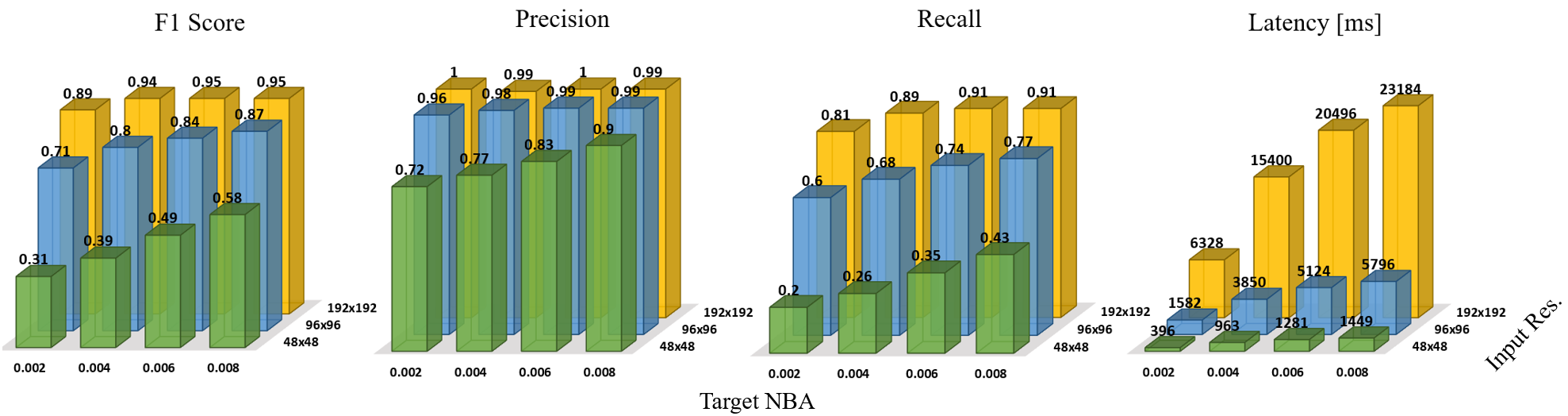}
\caption{This figure shows the resulting F1 metrics as well as the latency for different configurations of our method. The target \gls{nba}, indicated on the bottom axis determines the target object size that our tiling approach aims for.}
\label{fig:latency}
\vspace{-10pt}
\end{figure*}

\subsection{Experimental Settings}
As well as comparing the proposed method to other work on the CARPK \cite{carpk} dataset we also evaluate the performance gains, in terms of small object detection, that arise from different configurations of our tiling method. More specifically, we train models with three different input resolutions as well as four different values of target \gls{nba}. For this evaluation, we use the Sony \emph{Spresense}, a microcontroller equipped with 6 Cortex-M4F cores and 1.5 MB of SRAM. Our models are quantized using the TFLite library and we execute them on the device using the \gls{tflm} interpreter. Deploying our models to the \emph{Spresense} allows us to observe how\textemdash not only the detection performance\textemdash but also the memory consumption, and latency are affected by different target object sizes of the proposed tiling method as well as varying input resolutions to the network.


\section{Results}

\subsection{Carpark Monitoring}
\begin{figure}
\centering
\includegraphics[width=0.475\textwidth]{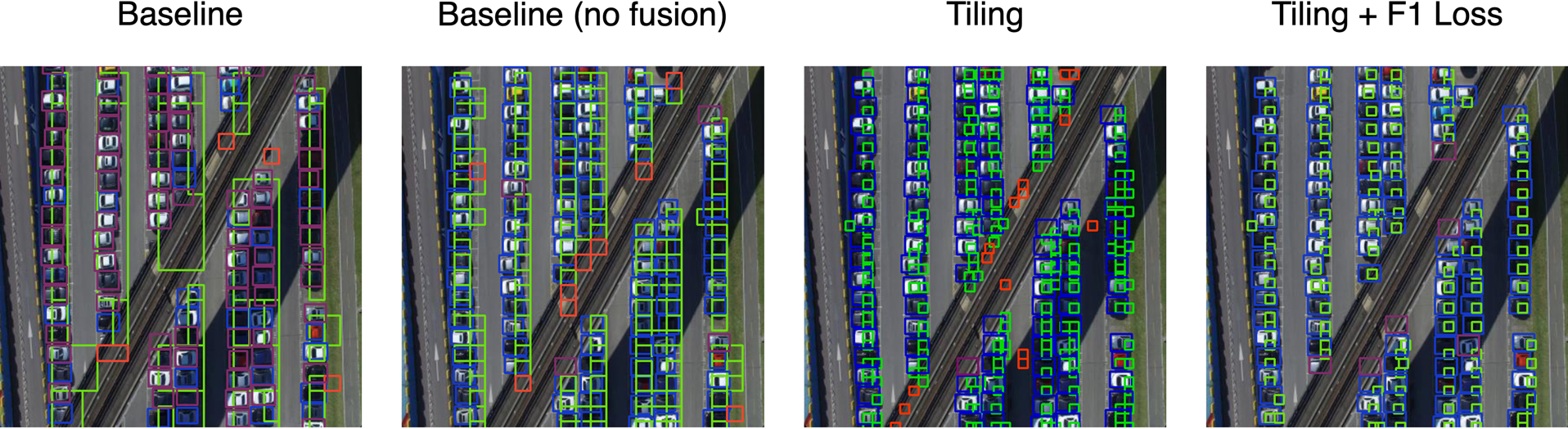}
\caption{The green rectangles indicate true positive predictions, the blue rectangles are ground truth object bounding boxes that were correctly predicted, the red rectangles show false positive predictions and lastly the purple rectangles are ground truth bounding boxes for cars that were not identified.}
\label{fig:ablation}
\vspace{-10pt}
\end{figure}

We evaluate our small object detection approach on the CARPK \cite{carpk} dataset. This dataset was collected by a drone flying at roughly 40 meters altitude over four different parking lots. It consists of close to 1500 images containing more than 90000 cars. In \cref{tab:count_error} we compare the object count error, the F1 score, and the model size of our method with the standard \gls{fomo} implementation as well as other published results on the CARPK dataset. Our approach achieves a reduction of $76\%$ in the object count error and a $225\%$ increase in F1 score compared to the original \gls{fomo} architecture. The \gls{mae} for \gls{yolo} is reported from \cite{carpk} and the value for \gls{yolo}v4 is reported from \cite{sf_ssd}. For both \gls{yolo} and \gls{yolo}v4 we estimate the number of parameters based on the implementations in \cite{yolo} and \cite{yolov4}, respectively. A visual demonstration of how the different components of our approach contribute to this improvement can be seen in \cref{fig:ablation}. Here it can be observed that removing the clustering step as well as adding the tiling drastically improves the recall. In the image on the far right in \cref{fig:ablation} we show how using the soft F1 loss improves the precision and count error by reducing the number of duplicate predictions on the same cars. Compared to \gls{yolo} \cite{yolo} our method reduces the \gls{mae} by $74\%$, while \gls{yolo}v4 \cite{yolov4} achieves a $85\%$ reduction. However, both \gls{yolo} and \gls{yolo}v4 are not designed for microcontrollers and do not fit the memory requirements of most low-power embedded devices. This is showcased clearly in \cref{tab:count_error} when comparing the number of model parameters.

\begin{table}
\centering
\caption[Comparison of Count Error on CARPK Data Set]{We compare the count error (MAE) of our method as well as the standard \gls{fomo} implementation with previous work on the CARPK data set.}
\begin{tabular}{l|l|l|l}
\toprule
\toprule
Method       & MAE           & F1   & Parameters\\
\midrule
\midrule
\gls{fomo}         & 54,32         & 0.28 & \textbf{19K}\\
\gls{yolo}         & 48,89         & - & 119M\\
\gls{yolo}v4       & \textbf{7,16}         & - & 35M\\
Ours         & 12,9          & \textbf{0.91} & \textbf{19K}\\
\bottomrule
\bottomrule
\end{tabular}

\label{tab:count_error}
\vspace{-10pt}
\end{table}

\subsection{On Device Performance}
In \cref{fig:latency} the F1 metrics as well as the on-device latency are shown for different configurations of our method. 
From the data in \cref{fig:latency} one can observe that both increasing the target \gls{nba}, i.e. using more tiles, as well as using a higher input resolution results in significantly improved F1 metrics. The strong improvement in detection performance suggests that the main obstacle in small object detection is indeed the relative lack of information that is available for small objects. These results also show that when increasing the input resolution is not an option\textemdash due to memory restrictions\textemdash the proposed tiling approach is a valid alternative to boost the signal-to-noise ratio of small objects and with that, the detection performance. What we can also see in these results is the additional latency that is incurred by the tiling method. For example, reaching the target \gls{nba} of 0.008 requires 3.66 times more tiles than the target \gls{nba} of 0.002, which results in a total latency increase of 16856 milliseconds to process an entire image, when using an input resolution of $192\times 192$.

\section{Conclusion}
We propose an adaptive tiling approach to overcome the challenges of small object detection on embedded devices. We demonstrate how our tiling method can be used to trade latency for high detection performance, rivaling state-of-the-art large-scale algorithms such as \gls{yolo}v4, while being able to fulfill the stringent low memory requirements of \gls{tinyml} on embedded devices. These findings show that our approach makes an important step towards enabling small object detection on embedded devices. Furthermore, we show that using a soft F1 loss can improve the balance between precision and recall in imbalanced datasets.

\section*{Acknowledgment}

The authors would like to thank Sony for collaborating and providing the computational resources, that were used to carry out all hardware evaluations for this work.

\newpage



%

\bibliographystyle{IEEEtran}
\bibliography{main.bib}



\end{document}